# Biometric Signature Processing & Recognition Using Radial Basis Function Network

Ankit Chadha, Neha Satam, and Vibha Wali

*Abstract*- Automatic recognition of signature is a challenging problem which has received much attention during recent years due to its many applications in different fields. Signature has been used for long time for verification and authentication purpose. Earlier methods were manual but nowadays they are getting digitized. This paper provides an efficient method to signature recognition using Radial Basis Function Network. The network is trained with sample images in database. Feature extraction is performed before using them for training. For testing purpose, an image is made to undergo rotation-translation-scaling correction and then given to network. The network successfully identifies the original image and gives correct output for stored database images also. The method provides recognition rate of approximately 80% for 200 samples.

*Keywords*- Database, Feature extraction, Radial Basis Function Network, Signature recognition.

## I. INTRODUCTION

In everyday life there are many places where we are needed to give identification to gain access. Its examples include internet account, credit cards, ATM machines, etc. We have to first provide our identification and then we can access them. These identifications are generally a login ID and an alphanumeric (maybe with special characters) password or a Personal Identification Number (PIN). Main drawback of this system is that, they provide 'verification' and not actually 'identification'. These two terms, though used interchangeably, are different. When a system 'verifies' a password, it just checks whether the password is right or wrong. It doesn't 'identify' which person is gaining access. Hence, anyone having knowledge of another person's password can easily get his/her data and can tamper it.

With the advent of new technology, these systems are being replaced by much more advanced techniques to 'identify' a person. These techniques are called biometrics, which involve checking a person's biological traits such as face, retina, fingerprint, iris, voice, signature etc. Formally, biometrics refers to the identification of humans by their characteristics or traits. The system already has the samples of characteristics of all users. When a person provides his/her identity, say fingerprint, system checks it with the records in database. When any of records match with the one being provided for authentication, the user is identified by system & given access.

Some of the applications of biometric recognition include driving licenses, immigration, national ID, passport, voter registration, security applications, medical records, personal device logon, desktop logon, human-robot interaction, human-computer interaction, smart cards etc.

There are certain biological measurements which qualify to be a biometric. As said above, face, retina, iris, signature are used as biometric characteristics because they are different for different people. Following are some requirements [1] which are needed to be satisfied in order to use them as biometric characteristics:

1. Universality: everyone should have the characteristic.

2. Uniqueness: two people should have some difference in characteristic.

3. Durability: the characteristic should remain same over the time.

4. Collectively: the characteristic should be available in quantity.

Though above are the properties of characteristics, practically, we have to consider some other issues also [1]:

1. Performance: the characteristics should give excellent speed and accuracy with minimum exploitation of resources.

2. Acceptability: people should accept the use of characteristic in their daily life.

3. Circumvention: represents how easily the system can be deceived.

A plain biometric system is equipped with sensory unit, a unit for feature extraction, a unit to perform the matching and decision-making unit. The sensory unit collects the biometric data of user. For the signature recognition purpose, a digital tablet along with pen-like stylus is provided which works as sensor. The prominent features of data are extracted using feature extraction unit. These features signify the original data but in using lesser attributes. They can be position and orientation of certain points, curved lines in signature, underlined alphabets etc. Then this feature set is matched against every set stored in the database and a matching score is generated. This score

Manuscript received September 30, 2013.
A. R. Chadha is with the Vidyalankar Institute of Technology, Mumbai, 400037, INDIA (phone: +91-9167884425; email:ankit.chadha@vit.edu.in).
N. S. Satam is with the Vidyalankar Institute of Technology, Mumbai, 400037, INDIA (phone: +91-9920395727; email:neha.satam@vit.edu.in).
V. S. Wali is an Assistant Professor in the Electronics and Telecommunications Engineering Department, Vidyalankar Institute of Technology, Mumbai, 400037, INDIA (e-mail: vibha.vali@vit.edu.in).

depends on how many points are matched between data in database and newly produced data. Once this score is obtained then it is responsibility of decision-making unit whether to accept the individual as identified user or not.

The age-old method for signature verification is manual one, in which a person manually checks the signature. If both the signatures are sufficiently similar, then the access is granted. An automated signature verification process will help improve the scenario. High-end biometric methods include iris, retina, face and fingerprint based identification and verification. Even though human features such as iris, retina and fingerprints are fixed for a person and the variations in the respective biometric attribute are low, special and relatively expensive hardware is needed for data acquisition in such systems. For the various purposes such as banking, financial transactions, document authentication, signature is used since long time. Also, the verification is relatively less expensive than other biometric systems [2].

Though the verification is efficient, it is difficult. There are some variations in a person's signature over the time, which needs to be accounted. Examples of the various variations observed in the signature of an individual have been illustrated in Fig. 1. Every time signature is made, it differs from previous one in some angle or scale or curving of letters.

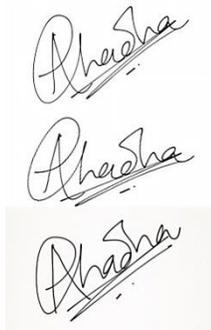

Fig 1: Variations in the signature of an individual

This involves eliminating or reducing the rotation, scaling and translation factors between the original and the newly produced signatures. For the verification & identification purpose, neural network can be employed. Fig.2 shows block diagram of proposed system, where all the effects of new signature are reduced and it is matched against those in database using neural network.

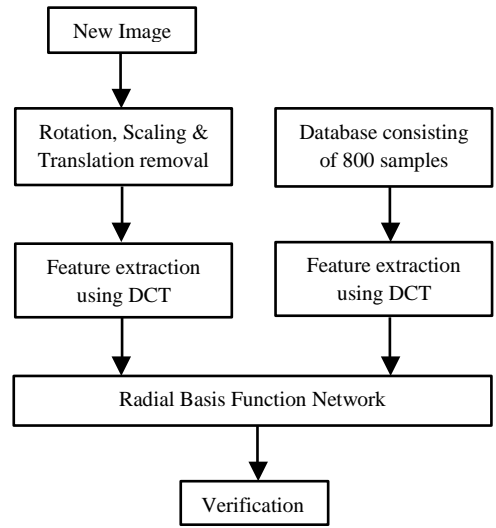

Fig.2: Block diagram of proposed system

First, new image, also known as user image is made to undergo Rotation-Scaling-Translation removal, so that it resembles original signature stored in database. Database is created by collecting 800 signature samples. These signatures are stored in database after feature extraction process for which DCT is employed. It is also applied on corrected user image. Now this image is provided to Radial Basis Function Network (RBFN), which is earlier trained using database. If RBFN identifies user image as the one in database then access for that particular individual is granted.

The advantage of this system is that it uses neural network effectively, i.e., it requires only few samples to train the network and then it performs for signatures newly added in the database. The algorithm used in this paper uses correlation to detect the rotation and implements simple cropping method to eliminate the effect. RBFN is used for verification purpose.

In paper [3], back-propagation algorithm is used for signature recognition. In this algorithm, neural network is created with R,G and B pixel values as input and gray value as output. To calculate the gradient, unsigned char to unsigned int and float to unsigned char conversions are needed to be performed. This increases computation. Also, the network is trained multiple times so that proper weight matrix is obtained. Use of RBFN reduces this need of multiple training and achieves the goal in few epochs.

This paper is organized as follows: Section II describes idea of proposed solution. Section III briefs implementation steps and section IV shows experimental results. Section V provides conclusion.

## II. IDEA OF PROPOSED SOLUTION

The primary concern is deciding at what angle the new signature deviates from the signature in database. For this, concept of correlation can be implemented. "Correlation" is a statistical measure, which refers to a process for establishing whether or not relationships exist between two variables [4]. It denotes the value by which two variable vary. If the value is maximum then there is maximum similarity in the two variables. Consider original signature as variable $P(m,n)$ and new signature to be tested as variable $Q(m,n)$. If $P_0$ and $Q_0$ are mean values of $P$ and $Q$ then cross-correlation $r$ between them is described by following equation:

$$r = \sum_m \sum_n (P_{mn} - P_0)(Q_{mn} - Q_0) \quad (1)$$

Normalized cross-correlation can be used to simplify examination and comparisons of coefficient values corresponding to the respective angular values. Min-max normalization is the procedure used to obtain normalized cross - correlation [5]. Min - max normalization maintains the relationships among the original data values. The normalization operation transforms the data into a new range, generally [0, 1]. Given a data set $x_i$, such that $i = 1, 2, \ldots n$, the normalized value $x'$ is given by the following equation:

$$x' = \frac{x - \min(x_i)}{\max(x_i) - \min(x_i)} \quad (2)$$

The second goal is to obtain the translation associated with the images. This can be accomplished by a simple cropping technique. Initially, the number of rows and columns bordering the signature pixels within the image are calculated. The image devoid of these rows and columns is extracted. The result is an image consisting of only the signature pixels. This eliminates additional background surrounding the image. Third goal is scaling between two images. For calculation of the scaling factor, the cropped images obtained during translation are used. The size of the original image divided by the size of the user image gives the scaling ratio.

*A. Sample Acquisition & Pre-processing of Images*

The system implemented here uses a digital pen tablet, namely, WACOM Bamboo shown in fig.3, as the data-capturing device. The pen has a touch sensitive switch in its tip such that only pen- down samples (i.e., when the pen touches the paper) are recorded.

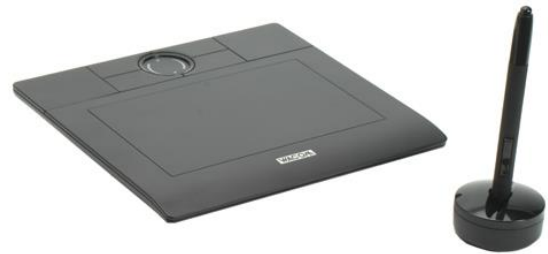

Fig. 3: Wacom Bamboo Digital Pen Tablet

For testing the system, a database was created. It consisted of a set of signature samples of 100 people. For each person, there are 9 test samples and 1 training sample. The samples are color normalized. Color normalization converts the colored image into corresponding grayscale image.

*B. Rotation Elimination*

There is always some tilting of signatures when there is no reference line provided for it. This angle of tilt varies from -60° to +60°. Before verification process, the rotation is neutralized by aligning the new image with original image. The new image is then rotated by 5° within the range of –60° to +60° in successive iterations. Cross-correlation values between the reference image and the new image are recorded on completion of each iteration of the rotation process. The maximum cross-correlation value refers to the correct angle of rotation within a 5° range, further, after the approximate angle value is obtained, +3° or –3° of this angle can be inspected for maximum correlation value which corresponds to angle of rotation accurate to up to 1°. The user image is rotated by the negative of the angle and we get the image which is free from rotation. Fig. 4 gives the flowchart of this algorithm.

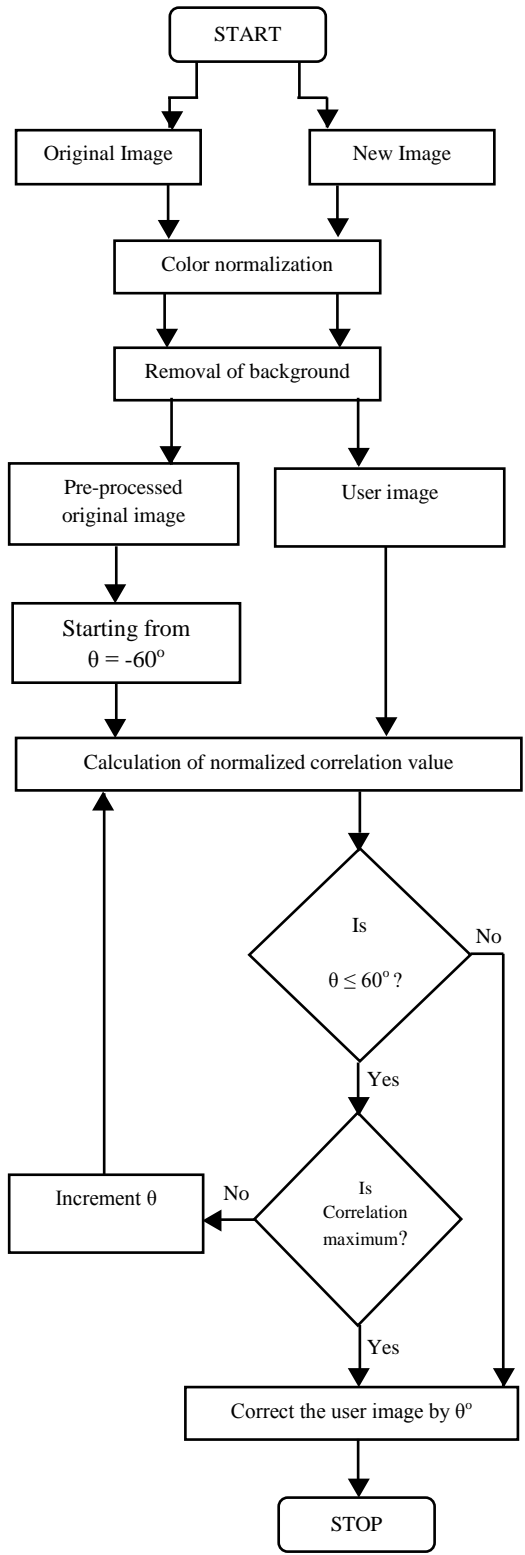

Fig.4: Flowchart of algorithm

### C. Translation Elimination

Similar to tilting of signature, there is one more difficulty which is translation. It is the effect which introduces translation in X and/or Y direction, having a maximum value equal to the width or height of the signature canvas respectively.

The solution is given by cropping method in which only the signature pixels are extracted. So with respect to new image formed, there is no such translation in X and/or Y direction. This eliminates the effect of translation. The number of columns from left and number of rows from the bottom, which contain no black pixels corresponding to the actual signature, i.e., which consist solely of image- background, are counted. These values give X translation and Y translation respectively.

### D. Scaling Elimination

The size of the signature varies according to the size of space. For smaller spaces, the signature may be compressed, for larger space, the sign may be enlarged. Thus, before extraction of features, it is essential that any scaling, if present in the test sample, be removed. Upon cropping both images, the ratio of height gives Y scaling and ratio of width gives X scaling. However, to resize the user image and make it the same size as the image, either of the scaling ratios can be used. For a rotation range of –60° to +60°, height was observed to vary significantly as compared to the length. Hence, Y scaling was chosen as the scaling ratio. Scaling ratio is calculated by the following equation:

$$\text{Scaling ratio} = \frac{\text{Size of reference image}}{\text{Size of test image}}$$

The user image is resized as per the obtained scaling ratio and then sent to the feature extraction segment.

### E. Combined Rotation–Scaling– Translation

It is not difficult to control the specimens to get pure rotation, translation and scaling, in any case, for original signature, all the aforementioned components are adjusted synchronously. Thus, rotation, translation and scaling modifications are connected in the same manner.

Rotation revision goes before translation revision as the expected cause at base left corner likewise gets rotated and translation impacts can't be dispensed with unless the origin is rotated back to bottom left as correctly as could be allowed. In this way, rotation adjustment needs to be performed first as the scaling proportion computed by the pure scaling strategy is not reliable with scaling degree of the rotated image.

Hence, the efficiency of scaling revision depends, to a substantial degree, on the percentage error obtained during rotation correction.

After correcting the angle of rotation, the user image preprocessed duplicate is edited to take out translation and the image so acquired is an instance of pure scaling which has been discussed previously. Along these lines, rotation–scaling–translation cancellation is performed.

*F. Feature Extraction using DCT*

DCT is one of the most widely used transform in the used for feature extraction. It involves taking the transformation of the image as a whole and separating the relevant coefficients. DCT performs energy compaction [6]. The DCT of an image basically consists of three frequency components namely low, middle, high each containing some detail and information in an image. The low frequency generally contains the average intensity of an image which is the most intended in FR systems [7]. Mathematically, the 2D-DCT of an image is given by:

$$F(u,v) = \alpha(u)\alpha(v) \sum_{x=0}^{N-1} \sum_{y=0}^{M-1} \cos\left[\frac{\pi u}{2N}(2x+1)\right] \cos\left[\frac{\pi u}{2M}(2y+1)\right] f(x,y) \quad (3)$$

$$\alpha(u)\alpha(v) = \begin{cases} \sqrt{\frac{1}{N}} \text{ for } u,v \neq 0 \\ \sqrt{\frac{2}{N}} \text{ for } u,v = 0 \end{cases}$$

where $f(x, y)$ is the intensity of the pixel at coordinates $(x, y)$, u varies from 0 to M-1, and v varies from 0 to N-1, where M × N is the size of image.

*G. Radial Basis Function Network (RBFN)*

The construction of RBFN in its basic from involves three layers, viz., input layer, hidden layer and output layer. All these layers are assigned different roles. The input layer is formed by sensory units that connect the network to its surrounding. The second layer, the hidden layer in entire network, applies nonlinear transformation from the input space to hidden space, where dimensionality of hidden space is larger than that of input space. Each neuron in hidden layer has a special type of activation function centered on the center vector of a cluster or subcluster in the feature space so that the function has non negligible response for input vectors close to its center. Output layer is linear, producing response of the network to the activation signals applied to input layer. This particular architecture of RBFN has proved to directly improve training and performance of the network [8]-[9]. Fig. 5 depicts architecture of RBFN.

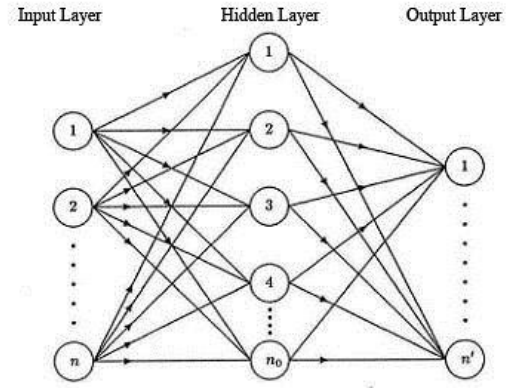

Fig. 5: General architecture of RBFN

The learning process undertaken by an RBFN can be described as follows. The linear weights associated with the output unit(s) of the network tend to evolve on a different time scale compared to nonlinear activation functions of the hidden unit. Thus as hidden layer's activation functions evolve slowly in accordance with some nonlinear optimization strategy; the output layer's weights adjust themselves rapidly through a linear optimization strategy. There are different learning strategies that can be followed, based on particular application. It depends on how centers of RBFN are specified. For this application of signature recognition, approach of supervised selection of centers is taken.

Several learning algorithms have been proposed in the literature for training RBF networks [10]–[15]. Selection of a learning algorithm for a particular application is critically dependent on its accuracy and speed. In practical online applications, sequential learning algorithms are generally preferred over batch learning algorithms as they do not require retraining whenever a new data is received.

In this approach, centers of RBFN and all other free parameters of the network undergo a supervised learning process. It means that RBFN takes its most generalized form of error-correction learning.

The first step in the development of such a learning procedure is to define the instantaneous value of the cost function [16].

$$\xi = \frac{1}{2} \sum_{j=1}^{N} e_j^2 \quad (3)$$

Where $N$ is the size of training sample used for learning and $e_j$ is the error signal defined by

$$e_j = d_j - F^*(x_j)$$

$$= d_j - \sum_{i=1}^{M} \omega_i G\left(\|x_j - t_i\|_{C_i}\right) \quad (4)$$

The requirement is to find the free parameters $\omega_i$, $t_i$ and $\sum_i^{-1}$ (the latter being related to norm-weighting matrix $C_i$) so as to minimize ξ.

### III. IMPLEMENTATION STEPS

#### A. Image Database

TABLE 1

COMBINED ROTATION–SCALING–TRANSLATION

| Sample | Original Parameters | | Detected Parameters | |
|---|---|---|---|---|
| | Rotation | Scaling | Rotation | Scaling |
| 1 | -50 | 0.54 | -48 | 0.58 |
| 2 | -20 | 0.9 | -22 | 0.98 |
| 3 | -10 | 0.65 | -8 | 0.68 |
| 4 | -5 | 1.43 | -3 | 1.47 |
| 5 | 3 | 1 | 1 | 1.2 |
| 6 | 12 | 0.83 | 10 | 0.88 |
| 7 | 15 | 1.5 | 17 | 1.6 |
| 8 | 32 | 1.78 | 30 | 1.84 |
| 9 | 33 | 0.75 | 35 | 0.79 |
| 10 | 42 | 0.26 | 40 | 0.32 |

For the experimental purpose, a database of 700 samples was created by collecting 10 samples each from 70 people. The network for identification was created by using RBFN fewer neurons model. All the images were used for training of the network. One of the images was processed by using rotation-translation-scaling algorithm. Fig.6 shows some of the signatures used in training and testing image database constructed.

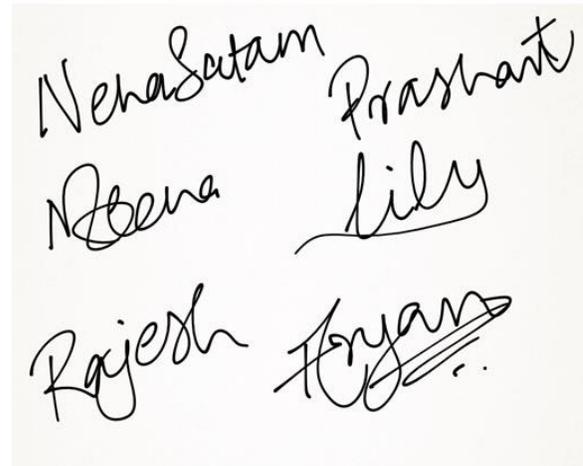

Fig.6: Some samples from signature database

The signature recognition system presented in this paper was developed, trained, and tested using MATLAB™ 7. The computer was a Windows 8 machine with a 2.5 GHz Intel Core I5 processor and 4 GB of RAM.

#### B. Validation of technique

The preprocessed grayscale images of size 8 × 8 pixels are reformed in MATLAB to form a 64 × 1 array with 64 rows and 1 column for each image. This technique is performed on all test images to form the input data for testing the recognition system. Similarly, the image database for training uses 50 images and forms a matrix of 64 × 50 with 64 rows and 50 columns. The input vectors defined for the RBFN are distributed over a 2D-input space varying over [0 255], which represents intensity levels of the grayscale pixels. As many as 5 test images are used with the image database for performing the experiments. Training and testing sets were used without any overlapping.

### IV. EXPERIMENTAL RESULTS

#### A. Combined Rotation–Scaling– Translation

Table 1 displays result for combined rotation–scaling–translation in original rotation and scaling and detected rotation and scaling.

TABLE 1

COMBINED ROTATION–SCALING– TRANSLATION

The graphical plot of error in actual and detected rotation angle is depicted in figure 7 below.

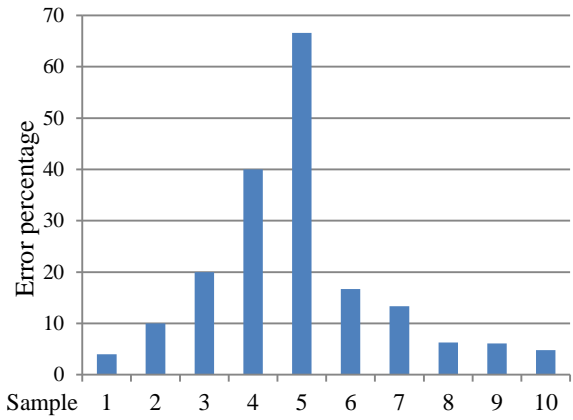

Fig. 7: Graphical plot of error in actual and detected rotation angle

The graphical plot of error in actual and detected scaling parameter can be seen in figure 8 below.

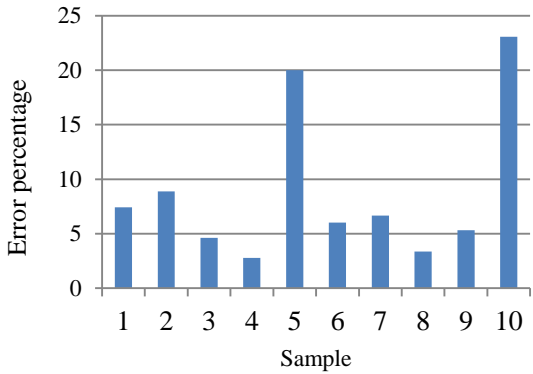

Fig. 8: Graphical plot of error in actual and detected scaling parameter

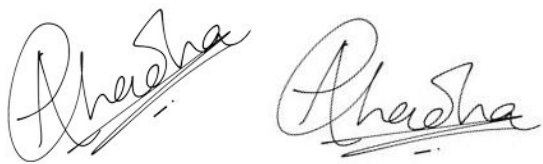

Original Image     Fig.10: Image after Rotation Correction    Fig.9:

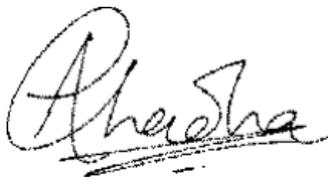

Fig.11: Image after Scaling Correction

### B. Recognition using MATLAB

The network was trained using variable number of samples. The performance graph of network for 700 samples is shown in figure 12.

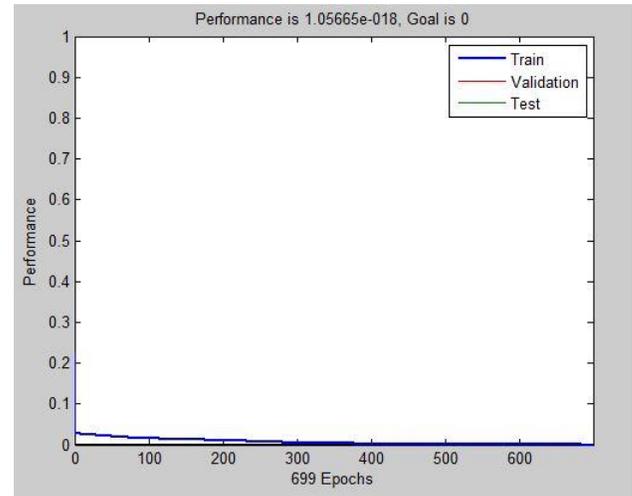

Fig. 12: Performance graph of network for 700 samples

The test image given to network is shown in figure 13. The network was successfully able to recognize it though the image was tilted.

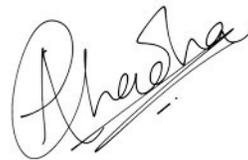 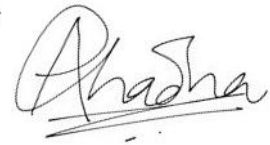

Fig. 13: User Image      Fig.14: Image recognized

Table 2 displays convergence rates relating to different Mean Square Errors (MSE) and spreads.

TABLE 2

CONVERGENCE RATES RELATING TO DIFFERENT MEAN SQUARE ERRORS (MSE) AND SPREADS

| MSE error | Spread | Iterations |
|---|---|---|
| 2.23695e-31 | 0.5 | 500 |
| 9.48458e-27 | 0.5 | 200 |
| 0.620313 | 0.5 | 100 |
| 2.41667 | 0.5 | 70 |
| 3.67695 | 0.5 | 60 |
| 5.25875 | 0.5 | 50 |
| 10.8547 | 0.5 | 40 |
| 15.4098 | 0.5 | 20 |
| 49.6859 | 0.5 | 10 |
| 211.032 | 0.5 | 5 |

For various numbers of samples in training set, the recognition rate of system also varied. It is tabulated in table.

TABLE 3

RECOGNITION RATE FOR VARIOUS NUMBERS OF SAMPLES

| Number of samples | Recognition rate |
|---|---|
| 500 | 71. 34% |
| 400 | 76.8% |
| 200 | 80% |
| 100 | 79.2% |
| 50 | 72.65% |

V. CONCLUSION

This paper presents a novel method to signature recognition using RBFN. The system was evaluated in MATLAB using an image database of 700 signatures, containing 70 people and each person with 10 signatures. After training for approximately 200 samples the system achieved a recognition rate of 80%. A reduced feature space substantially reduces the computational requirements of the method as compared with standard DCT feature extraction methods. This makes our system well suited for low-cost, real-time hardware implementation. Commercial implementations of this technique do not currently exist. However, it is conceivable that a practical RBFN-based signature recognition system may be possible in the future.

VI. FUTURE WORK

The system is image-based and can be extended to real time signature recognition also. With IOS and Android based systems, touch screen devices like tablets, phablets and smartphones can be effectively used for digital signature identification and authentication. Genetic algorithm can be utilized to solve this problem of verification with reduced computation and increased efficiency.

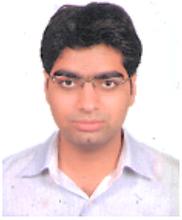 Ankit R. Chadha (M'2010) was born in Mumbai (M.H.) in India on November 07, 1992. He is currently pursuing his undergraduate studies in the Electronics and Telecommunication Engineering discipline at Vidyalankar Institute of Technology Mumbai. His special fields of interest include Image Processing, Computer Vision (particularly, Pattern Recognition) and Embedded Systems. He has 6 papers in International Journals & Conferences to his credit.

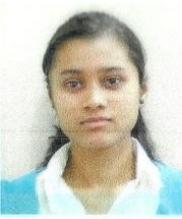 Neha S. Satam (M'2010) was born in Mumbai (M.H.) in India on November 26, 1992. She is currently pursuing her undergraduate studies in the Electronics and Telecommunication Engineering discipline at Vidyalankar Institute of Technology, Mumbai. Her fields of interest include Image Processing, Stegnography and Embedded systems. She has 6 papers in International Journals & Conferences to her credit.

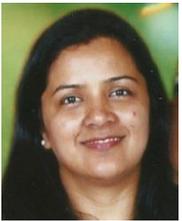 Prof. Vibha S. Wali (M'1997) has done B.E. in Electronics in 1997 from Shivaji University and Masters in Electronics and Telecommunication from Mumbai University. She had taught subjects like computer graphics, object oriented programming, data structure, Image processing and many more. She is currently working as Assistant professor in Vidyalankar college of Technology, Wadala in Information Technology department and presently teaching subjects like Electromagnetic Wave Theory, Electromagnetic Engineering, Neural network and Fuzzy Logic and Radio Frequency Circuit Design. Her area of interest is Image processing and Fractal Image compression. She has guided numerous undergraduate projects and papers.